\documentclass{article}
\usepackage{ijcai22}

\usepackage{times}
\usepackage{soul}
\usepackage{url}
\usepackage[hidelinks]{hyperref}
\usepackage[utf8]{inputenc}
\usepackage[small]{caption}
\usepackage{graphicx}
\usepackage{amsmath}
\usepackage{amsthm}
\usepackage{booktabs}
\usepackage{algorithm}
\usepackage{algorithmic}
\usepackage{multirow}
\usepackage{xspace}
\usepackage{amsmath}
\usepackage{amsthm}
\usepackage{multirow}
\usepackage{graphicx}
\usepackage{multirow}
\usepackage{booktabs}
\usepackage{amssymb}
\usepackage{stmaryrd}
\usepackage{bbm}
\usepackage{mathtools}
\usepackage{cancel}
\usepackage{caption}
\usepackage{subcaption}
\usepackage{algorithm}
\newcommand{\entropy}{\textsc{Ent}}
\newcommand{\cache}{\mathsf{c}}
\usepackage[usenames,dvipsnames]{xcolor}
\usepackage{footnote}

\newcommand{\LineIf}[2]{     
    \STATE \algorithmicif\ {#1}\ \algorithmicthen\ {#2} 
}

\theoremstyle{definition}

\newcommand{\id}[1]{\llbracket{#1}\rrbracket}
\newcommand{\Ind}[1]{{\ensuremath{\mathbbm{1}\!\left\{#1\right\}}}}

\usepackage{siunitx,etoolbox}
\newrobustcmd{\B}{\bfseries}
\title{Neuro-Symbolic Entropy Regularization %
}

\author{
Kareem Ahmed\and
Eric Wang\and
Kai-Wei Chang\And
Guy Van den Broeck\\
\affiliations
Computer Science Department\\
University of California, Los Angeles\\
Los Angeles, CA, USA\\
\emails
ahmedk@cs.ucla.edu,
ericzxwang@ucla.edu,
\{kwchang, guyvdb\}@cs.ucla.edu
}

\newcommand{\rvars}[1]{\ensuremath{\mathbf{#1}}\xspace}
\newcommand{\Xs}{\rvars{X}}
\newcommand{\Ys}{\rvars{Y}}

\newcommand{\jstate}[1]{\ensuremath{\mathbf{#1}}\xspace}
\newcommand{\xs}{\jstate{x}}
\newcommand{\ys}{\jstate{y}}

\newcommand{\vect}[1]{\ensuremath{\mathbf{\mathsf{#1}}}\xspace}
\newcommand{\pv}{\vect{p}}

\newcommand{\ch}{\ensuremath{\mathsf{in}}}
\newcommand{\vars}{\ensuremath{\mathsf{vars}}}

\newcommand{\Y}{\ensuremath{\mathbf{Y}}}
\newcommand{\y}{\ensuremath{\mathbf{y}}}



\newcommand{\ra}[1]{\renewcommand{\arraystretch}{#1}}

\newif\ifcomments
\ifcomments
    \providecommand{\kareem}[2][]{{\protect\color{red}{[Kareem:\textbf{#1} #2]}}}
    \providecommand{\kaiwei}[2][]{{\protect\color{red}{[Kaiwei:\textbf{#1} #2]}}}
    \providecommand{\guy}[2][]{{\protect\color{purple}{[Guy:\textbf{#1} #2]}}}
    \providecommand{\eric}[2][]{{\protect\color{ForestGreen}{[Eric:\textbf{#1} #2]}}}
\else
    \providecommand{\kareem}[2][]{}
    \providecommand{\kaiwei}[2][]{}
    \providecommand{\guy}[2][]{}
    \providecommand{\eric}[2][]{}
\fi

\usepackage{tikz}
\usetikzlibrary{circuits.logic.US}
\usetikzlibrary{shapes.misc}
\usetikzlibrary{shapes.arrows}
\usetikzlibrary{arrows,shapes.geometric}
\usetikzlibrary{positioning}

\usepackage{pgfplots}
\pgfmathdeclarefunction{gaussian}{2}{%
  \pgfmathparse{1/(#2*sqrt(2*pi))*exp(-((x-#1)^2)/(2*#2^2))}%
}
\pgfmathdeclarefunction{gaussianmixture2}{5}{%
  \pgfmathparse{#5*(1/(#2*sqrt(2*pi))*exp(-((x-#1)^2)/(2*#2^2))) + (1-#5)*(1/(#4*sqrt(2*pi))*exp(-((x-#3)^2)/(2*#4^2)))}%
}

\newcommand{\figref}[1]{Fig.~\ref{#1}}

\newcommand{\citet}[1]{\citeauthor{#1}~[\citeyear{#1}]}
\newlength{\picHeight}

\begin{document}

\maketitle
\begin{abstract}
In structured prediction, the goal is to jointly predict many output variables that together encode a structured object -- a path in a graph, an entity-relation triple, or an ordering of objects.
Such a large output space makes learning hard and requires vast amounts of labeled data.
Different approaches leverage alternate sources of supervision.
One approach -- entropy regularization -- posits that 
decision boundaries should lie in low-probability regions.
It extracts supervision from unlabeled examples, but remains agnostic to the structure of the output space.
Conversely, neuro-symbolic approaches exploit the knowledge that not every prediction corresponds to a \emph{valid} structure in the output space. Yet, they does not further restrict the learned output distribution.
This paper introduces a framework that unifies both approaches.
We propose a loss, \emph{neuro-symbolic entropy regularization}, that encourages the model to confidently predict a valid object.
It is obtained by restricting entropy regularization to the distribution over only valid structures.
This loss is efficiently computed when the output constraint is expressed as a tractable logic circuit.
Moreover, it seamlessly integrates with other neuro-symbolic losses that eliminate invalid predictions.
We demonstrate the efficacy of our approach on a series of semi-supervised and fully-supervised structured-prediction experiments, where we find that it leads to models whose predictions are more accurate and more likely to be valid.

\end{abstract}

\section{Introduction}

Neural networks have achieved breakthroughs across a wide range of domains.
Such breakthroughs are often only possible in the presence of large labeled datasets, which can be hard to obtain.
Increasing efforts are therefore being devoted to approaches that utilize alternate sources of supervision in lieu of \emph{more} labeled data. 
Entropy regularization constitutes one such approach~\cite{grandvalet2005,ssl}.
It posits that data belonging to the same class tend to form discrete clusters.
Minimizing the entropy of the predictive distribution can thus be regarded as minimizing a measure of class overlap under the learned representation.
Intuitively, a classifier guessing uniformly at random has \emph{maximum entropy}, and has not learned features informative of the underlying class.
Consequently, we prefer a \emph{minimum entropy} classifier that learns features \emph{maximally informative} of the underlying class, even on unlabeled data.

The need for labeled data is only exacerbated in structured prediction, where the objective is to predict multiple interdependent output variables representing a discrete object.
Viewed as traditional classification, the number of classes in structured prediction is exponential in the number of output variables -- all possible output configurations.
Neuro-symbolic methods can provide additional supervision leveraging symbolic knowledge regarding the structure of the output space~\cite{RaedtIJCAI2020}.
This knowledge, typically expressed in logic, characterizes the set of valid structures; for instance, not every selection of edges in a graph is a path.

In this paper we take a principled approach to unifying the aforementioned forms of supervision.
Naively, we might consider simply optimizing both losses simultaneously. 
However, computed in that manner, the entropy does not account for the output-space structure, and is therefore likely to push the network towards invalid structures.
Instead, we restrict the entropy loss to the network's distribution over the valid structures, as characterized by the constraint, as opposed to the entire predictive distribution, proposing a new loss \emph{neuro-symbolic entropy regularization}.
That is, we require that the network's output distribution be maximally informative \emph{subject to the constraint}.
Intuitively, the network should ``know'' the right structure among the valid structures.
Computing the entropy of a distribution subject to a constraint is, in general, computationally hard.
We provide an algorithm leveraging structural properties of tractable logical circuits to efficiently compute this quantity.
Our framework integrates seamlessly with other neuro-symbolic approaches that maximize the constraint probability, in effect ``eliminating'' invalid structures.

Empirically, we evaluate our loss on four structured prediction tasks, in both semi-supervised and fully-supervised settings. We observe it leads to models whose predictions are more accurate, \emph{as well as} more likely to satisfy the constraint.

\section{Neuro-Symbolic Entropy Loss}\label{sec:nser}

We first introduce background on logical constraints and probability distributions over output structures.
Afterwards, we motivate and define our neuro-symbolic entropy loss.

\subsection{Background}\label{sec:background}
We write uppercase letters ($X$, $Y$) for Boolean variables and lowercase letters ($x$, $y$) for their instantiation ($Y=0$ or $Y=1$).
Sets of variables are written in bold uppercase ($\Xs$, $\Ys$), and their joint instantiation in bold lowercase ($\xs$, $\ys$).
A literal is a variable ($Y$) or its negation ($\neg Y$).
A logical sentence ($\alpha$ or $\beta$) is constructed from variables and logical connectives ($\land$, $\lor$, etc.), and is also called a (logical) formula or constraint.
A state or world $\ys$ is an instantiation to all variables $\Ys$.
A state $\ys$ satisfies a sentence $\alpha$, denoted $\ys \models \alpha$, if the sentence evaluates to true in that world. A state $\ys$ that satisfies a sentence $\alpha$ is also said to be a model of $\alpha$.
We denote by $m(\alpha)$ the set of all models of $\alpha$
The notation for states $\ys$ is used to refer to an assignment, the logical sentence enforcing the assignment, or the binary output vector capturing the assignment, as these are all equivalent notions.
A sentence $\alpha$ entails another sentence $\beta$, denoted $\alpha \models \beta$, if all worlds that satisfy $\alpha$ also satisfy $\beta$.

\paragraph{A Probability Distribution over Possible Structures}
Let $\alpha$ be a logical sentence defined over Boolean variables $\Ys = \{Y_1,\dots,Y_n\}$.
Let $\pv$ be a vector of probabilities for the same variables $\Ys$, where $\pv_i$ denotes the predicted probability of variable $Y_i$ and corresponds to a single output of the neural network.
The neural network's outputs induce a probability distribution $\Pr(\cdot)$ over all possible states $\ys$ of $\alpha$
\begin{equation}\label{eqn:pr_struct}
     \Pr(\y) = \prod_{i: \y \models \Y_i} \pv_i \prod_{i: \y \models \lnot \Y_i} (1 - \pv_i).
\end{equation}

\paragraph{Semantic Loss}\label{sec:semantic_loss}
The semantic loss is a function of $\alpha$ and $\pv$. 
It quantifies how close the neural network comes to satisfying the constraint by computing the probability of the constraint under the distribution $\Pr(\cdot)$.
It does so by reducing the problem of probability computation to the weighted model counting (WMC): summing up the models of $\alpha$, each weighted by its likelihood under $\Pr(\cdot)$.
It, therefore, maximizes the probability mass allocated by the network to the models of $\alpha$
\begin{equation}
\label{eq:sloss}
\mathbb{E}_{\y \sim \Pr}\left[ \Ind{\y \models \alpha} \right] 
= \sum_{\y \models \alpha} \Pr(\y).
\end{equation}
Taking the negative logarithm recovers semantic loss. We use semantic loss in experiments to "eliminate" invalid structures.

\subsection{Motivation and Definition}

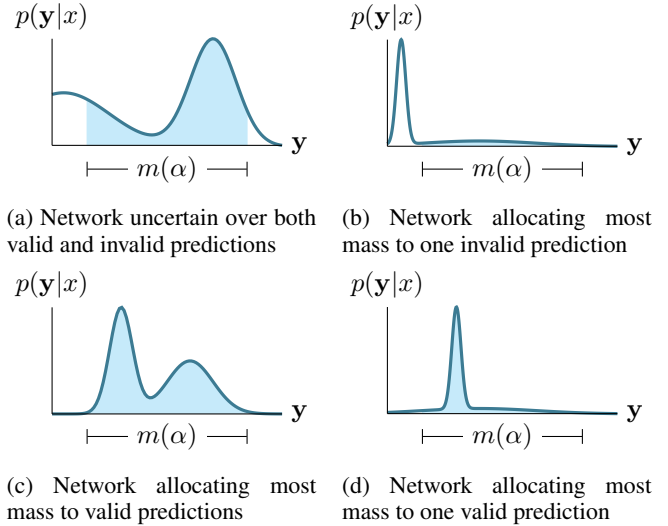
\begin{figure}[t]
\begin{subfigure}[b]{0.48\columnwidth}
\centering
\begin{tikzpicture}
\begin{axis}[
    no markers, domain=0:10, samples=100,
    axis lines*=left, xlabel=$\y$, ylabel=$p(\y|x)$,
    every axis y label/.style={at=(current axis.above origin),anchor=south},
    every axis x label/.style={at=(current axis.right of origin),anchor=west},
    height=3cm,
    width=132pt,
    xtick=\empty, ytick=\empty,
    enlargelimits=false, clip=false, axis on top,
    grid = major,
    ]
    \addplot [fill=cyan!20, draw=none, domain=1.5:8.5] {gaussianmixture2(0.5,2,7,1,0.5)} \closedcycle;
    \addplot [very thick,cyan!50!black] {gaussianmixture2(0.5,2,7,1,0.5)};
    \draw [yshift=-0.3cm, latex-latex, |-|](axis cs:1.5,0) -- node [fill=white] {$m(\alpha)$} (axis cs:8.5,0);
\end{axis}
\end{tikzpicture}
\caption{Network uncertain over both valid and invalid predictions}
\end{subfigure}
\hfill
\begin{subfigure}[b]{0.48\columnwidth}
\centering
\begin{tikzpicture}
\begin{axis}[
    no markers, domain=0:10, samples=500,
    axis lines*=left, xlabel=$\y$, ylabel=$p(\y|x)$,
    every axis y label/.style={at=(current axis.above origin),anchor=south},
    every axis x label/.style={at=(current axis.right of origin),anchor=west},
    height=3cm,
    width=132pt,
    xtick=\empty, ytick=\empty,
    enlargelimits=false, clip=false, axis on top,
    grid = major,
    ]
    \addplot [fill=cyan!20, draw=none, domain=1.5:8.5] {gaussianmixture2(0.6,0.2,4,2.5,0.6)} \closedcycle;
    \addplot [very thick,cyan!50!black] {gaussianmixture2(0.6,0.2,4,2.5,0.6)};
    \draw [yshift=-0.3cm, latex-latex, |-|](axis cs:1.5,0) -- node [fill=white] {$m(\alpha)$} (axis cs:8.5,0);
\end{axis}
\end{tikzpicture}
\caption{Network allocating most mass to one invalid prediction}
\end{subfigure}

\begin{subfigure}[b]{0.48\columnwidth}
\centering
\begin{tikzpicture}
\begin{axis}[
    no markers, domain=0:10, samples=100,
    axis lines*=left, xlabel=$\y$, ylabel=$p(\y|x)$,
    every axis y label/.style={at=(current axis.above origin),anchor=south},
    every axis x label/.style={at=(current axis.right of origin),anchor=west},
    height=3cm,
    width=132pt,
    xtick=\empty, ytick=\empty,
    enlargelimits=false, clip=false, axis on top,
    grid = major,
    ]
    \addplot [fill=cyan!20, draw=none, domain=1.5:8.5] {gaussianmixture2(3,0.5,6,1,0.5)} \closedcycle;
    \addplot [very thick,cyan!50!black] {gaussianmixture2(3,0.5,6,1,0.5)};
    \draw [yshift=-0.3cm, latex-latex, |-|](axis cs:1.5,0) -- node [fill=white] {$m(\alpha)$} (axis cs:8.5,0);
\end{axis}
\end{tikzpicture}
\caption{Network allocating most mass to valid predictions}
\end{subfigure}
\hfill
\begin{subfigure}[b]{0.48\columnwidth}
\centering
\begin{tikzpicture}
\begin{axis}[
    no markers, domain=0:10, samples=500,
    axis lines*=left, xlabel=$\y$, ylabel=$p(\y|x)$,
    every axis y label/.style={at=(current axis.above origin),anchor=south},
    every axis x label/.style={at=(current axis.right of origin),anchor=west},
    height=3cm,
    width=132pt,
    xtick=\empty, ytick=\empty,
    enlargelimits=false, clip=false, axis on top,
    grid = major,
    ]
    \addplot [fill=cyan!20, draw=none, domain=1.5:8.5] {gaussianmixture2(3,0.2,4,2.5,0.6)} \closedcycle;
    \addplot [very thick,cyan!50!black] {gaussianmixture2(3,0.2,4,2.5,0.6)};
    \draw [yshift=-0.3cm, latex-latex, |-|](axis cs:1.5,0) -- node [fill=white] {$m(\alpha)$} (axis cs:8.5,0);
\end{axis}
\end{tikzpicture}
\caption{Network allocating most mass to one valid prediction}
\end{subfigure}
\caption{%
A network's predictive distribution can be uncertain or certain ($\leftrightarrow$), and it can allow or disallow invalid predictions under the constraint $\alpha$ ($\updownarrow$).
Entropy regularization steers the network towards confident, possibly invalid predictions (b). 
Neuro-symbolic learning steers the network towards valid predictions without necessarily being confident (c).
Neuro-symbolic entropy-regularization guides the network to valid and confident predictions~(d).
}
\label{fig:entsl}
\end{figure}

Consider the plots in Figure \ref{fig:entsl}.
A neural network can be fairly uncertain regarding the target class accommodating of both valid and invalid predictions under its learned distribution.

A common underlying assumption in many machine learning methods is that data belonging to the same class tend to form discrete clusters\;\cite{ssl} -- an assumption deemed justified on the sheer basis of the existence of classes.
Consequently, a classifier is expected to favor decision boundaries lying in regions of low data density, separating the clusters.
Entropy-regularization \cite{grandvalet2005} directly implements the above assumption, requiring the classifier output confident -- low-entropy -- predictive distributions , pushing the decision boundary away from unlabeled points, thereby supplementing scarce labeled data with abundant unlabeled data.
Through that lens, minimizing the entropy of the predictive distribution can be seen as minimizing a measure of class overlap under the learned features.

Entropy regularization, however, fails to exploit situations where we have knowledge characterizing valid predictions in the domain.  
It can often be detrimental to the model's performance by guiding it towards confident yet invalid predictions.

Conversely, neuro-symbolic approaches steer the network towards distributions disallowing invalid predictions, by maximizing the constraint probability, but do little by way of ensuring the network learn features conducive to classification.

Clearly then, there is a benefit to combining the merits of both approaches. We restrict the entropy computation to the distribution over models of the logical formula, ensuring the network only grow confident in valid predictions. Complemented with maximizing the constraint probability, the network learns to allocate all of its mass to models of the constraint, while being maximally informative of the target.

\paragraph{Defining the Loss}
More precisely, let $\Y$ be a random variable distributed according to eqn. \eqref{eqn:pr_struct}, $\Y\sim \Pr(\cdot)$. We are interested in minimizing the entropy of $\Y$ conditioned on $\alpha$
\begin{equation}
    \begin{aligned}\label{eqn:nsentropy}
    H(\Y | \alpha)  &= - \sum_{\y \models \alpha} \Pr(\y | \alpha) \log \Pr(\y | \alpha)\\
                    &= - \mathbb{E}_{\Y | \alpha} \left[ \log \Pr(\Y | \alpha) \right].
    \end{aligned}
\end{equation}

\section{Computing the Loss}\label{sec:compute_nser}
The above loss is, in general, hard to compute. 
To see this, consider the uniform distribution over models of a constraint $\alpha$.
That is, let $\Pr(\y|\alpha) = \frac{1}{|m(\alpha)|}$ for all $\y \models \alpha$.
Then, $H(\Y | \alpha) = -\sum_{\y \models \alpha} \frac{1}{|m(\alpha)|} \log \frac{1}{|m(\alpha)|} = \log |m(\alpha)|$.
This tells us how many models of $\alpha$ there are, which is a well-known \#P-hard problem \cite{Valiant1979a,Valiant1979b}.
We will show that, through compilation into tractable circuits, we can compute eqn. \eqref{eqn:nsentropy} in time linear in the size of the circuit.

\begin{algorithm}[tb]
   \caption{\textsc{Ent}($\alpha, \Pr, \cache$)}
   \label{alg:Shannon-Entropy}
   {\bfseries Input:} a smooth, deterministic and decomposable logical circuit $\alpha$, a fully-factorized probability distribution $\Pr(\cdot)$ over states of $\alpha$, and a cache $\cache$ for memoization\\
   {\bfseries Output:} $H_{\Pr}(\Y | \alpha)$, where $\Y \sim \Pr(\cdot)$
   
   \begin{algorithmic}[1]
   \LineIf{$\alpha \in \mathsf{c}$}{\textbf{return} $\cache(\alpha)$}
   \IF{$\alpha$ is a literal}
   \STATE $e \leftarrow 0$
   \ELSIF{$\alpha$ is an AND gate}
   \STATE $e \leftarrow \entropy(\beta, \Pr, \cache) + \entropy(\gamma, \Pr, \cache)$
   \ELSIF{$\alpha$ is an OR gate}
   \STATE \parbox[t]{1.1\linewidth}{$e\leftarrow\sum_{i=1}^{|\ch(\alpha)|}\!\Pr(\beta_i) \log \Pr(\beta_i) \! + \! \Pr(\beta_i) \,  \entropy(\beta_i, \Pr, \cache)$}
   \ENDIF
   \STATE $\mathsf{c}(\alpha)\leftarrow e$
   \STATE \textbf{return} $e$
\end{algorithmic}
\end{algorithm}
\subsection{Computation through Compilation}
\paragraph{Tractable Circuit Compilation}
We resort to knowledge compilation techniques -- a class of methods that transform, or \emph{compile}, a logical theory into a target form with certain properties that allow certain probabilistic queries to be answered efficiently.
More precisely, we know of circuit languages that compute the probability of constraints~\cite{darwiche03}, and that are amenable to backpropagation.
We use the circuit compilation techniques in~\citet{darwiche11} to build a logical circuit representing our constraint.
Due to the structural properties of this circuit form, we can use it to compute both the probability of the constraint as well as its gradients with respect to the network's weights, in time linear in the size of the circuit~\cite{darwiche02}. 
This does not, in general, escape the complexity of the computation: worst case, the compiled circuit can be exponential in the size of the constraint. In practice, however, constraints often exhibit enough structure (repeated sub-problems) to make compilation feasible.
We refer to the literature for details of this compilation.

\paragraph{Logical Circuits} 
More formally, a \emph{logical circuit} is a directed, acyclic computational graph representing a logical formula.
Each node $n$ in the DAG encodes a logical sub-formula, denoted $[n]$.
Each inner node in the graph is either an AND or an OR gate, and each leaf node encodes a Boolean literal ($Y$ or $\lnot Y$). 
We denote by $\ch(n)$ the set of $n$'s children.

\paragraph{Structural Properties}  As already alluded to, circuits enable the tractable computation of certain classes of queries over encoded functions granted that a set of structural properties are enforced. We explicate such properties down below.

A circuit is \emph{decomposable} if the inputs of every AND gate depend on disjoint sets of variables i.e. for $\alpha = \beta \land \gamma$, $\vars(\beta) \cap \vars(\gamma) = \varnothing$.
Intuitively, decomposable AND nodes encode local factorizations of the function. For simplicity, we assume decomposable AND gates to have two inputs, a condition enforceable on any circuit for a polynomial increase in its size~\cite{vergari2015simplifying,peharz2020einsum}.

A second useful property is \emph{smoothness}.
A circuit is \emph{smooth} if the children
of every OR gate depend on the same set of variables i.e. for $\alpha = \bigvee_i \beta_i, \vars(\beta_i) = \vars(\beta_j)\ \forall i,j$. Decomposability and smoothness are a sufficient and necessary condition for tractable integration over arbitrary sets of variables in a single pass, as they allow larger integrals to decompose into smaller ones~\cite{choi2020pc}.

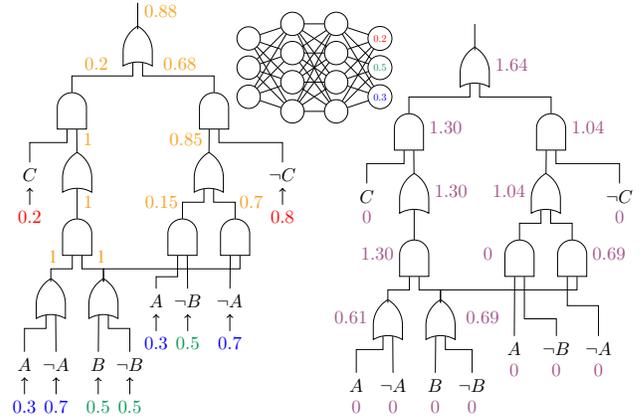
\begin{figure}[t!]
 \begin{subfigure}[b]{0.35\columnwidth}
 \centering
\scalebox{0.7}{
\begin{tikzpicture}[circuit logic US]
\node (output) at (4.25, 8) {};
\node (or1) [or gate, inputs=nn, rotate=90, scale=0.9] at (4.25,7) {};
\draw (or1) -- (output) node[pos=0.2, above right, color=YellowOrange] {$0.88$};

\node (and1) [and gate, inputs=nn, rotate=90, scale=0.9] at (3,5.8) {};
\node (and2) [and gate, inputs=nn, rotate=90, scale=0.9] at (5.7,5.8) {};

\node (c) at (2.2,4.6) {$C$};
\node (cval) at (2.2,3.8) {$\color{red}{0.2}$};
\draw[->] (cval) edge (c);
\node (nc) at (7,4.6) {$\neg C$};
\node (ncval) at (7,3.8) {$\color{red}0.8$};
\draw[->] (ncval) edge (nc);
\node (or2) [or gate, inputs=nnn, rotate=90, scale=0.9] at (3.1,4.6) {};
\node (or3) [or gate, inputs=nn, rotate=90, scale=0.9] at (5.6,4.6) {};

\node (and3) [and gate, inputs=nn, rotate=90, scale=0.9] at (3.1,3.4) {};
\node (and4) [and gate, inputs=nn, rotate=90, scale=0.9] at (5.1,3.4) {};
\node (and5) [and gate, inputs=nn, rotate=90, scale=0.9] at (6.1,3.4) {};

\node (a) at (4.6,2.2) {$A$};
\node (aval) at (4.6,1.4) {$\color{blue}0.3$};
\draw[->] (aval) edge (a);
\node (nb) at (5.2,2.2) {$\neg B$};
\node (nbval) at (5.2,1.4) {$\color{ForestGreen}0.5$};
\draw[->] (nbval) edge (nb);
\node (na) at (6,2.2) {$\neg A$};
\node (naval) at (6,1.4) {$\color{blue}0.7$};
\draw[->] (naval) edge (na);

\node (or4) [or gate, inputs=nn, rotate=90, scale=0.9] at (2.6,2.2) {};
\node (or5) [or gate, inputs=nn, rotate=90, scale=0.9] at (3.6,2.2) {};
\node (a1) at (2.1,1) {$A$};
\node (a1val) at (2.1,0.2) {$\color{blue}0.3$};
\draw[->] (a1val) edge (a1);
\node (na1) at (2.7,1) {$\neg A$};
\node (na1val) at (2.7,0.2) {$\color{blue}0.7$};
\draw[->] (na1val) edge (na1);
\node (b) at (3.5,1) {$B$};
\node (bval) at (3.5,0.2) {$\color{ForestGreen}0.5$};
\draw[->] (bval) edge (b);
\node (nb1) at (4.1,1) {$\neg B$};
\node (nb1val) at (4.1,0.2) {$\color{ForestGreen}0.5$};
\draw[->] (nb1val) edge (nb1);

\draw (or1.input 1) -- ++ (down: 0.25) -| (and1) node[pos=0.45, above right, color=YellowOrange] {$0.2$};
\draw (or1.input 2) -- ++ (down: 0.25) -| (and2) node[pos=0.1, above right, color=YellowOrange] {$0.68$};

\draw (and1.input 1) -- ++ (down: 0.25) -| (c);
\draw (and1.input 2) -- (or2) node[pos=0.45, right, color=YellowOrange] {$1$};

\draw (and2.input 2) -- ++ (down: 0.25) -| (nc);
\draw (and2.input 1) -- (or3) node[pos=0.45, left, color=YellowOrange] {$0.85$};

\draw (or2.input 2) -- (and3) node[pos=0.45, right, color=YellowOrange] {$1$};

\draw (or3.input 1) -- ++ (down: 0.25) -| (and4) node[pos=0.45, left, color=YellowOrange] {$0.15$};
\draw (or3.input 2) -- ++ (down: 0.25) -| (and5) node[pos=0.45, right, color=YellowOrange] {$0.7$};

\draw (and3.input 1) -- ++ (down: 0.25) -| (or4) node[pos=0.45, above, color=YellowOrange] {1};
\draw (and3.input 2) -- ++ (down: 0.25) -| (or5) node[pos=0.45, above, color=YellowOrange] {1};

\draw (and4.input 1) -- ++ (down:0.4) -| (a);
\draw (and4.input 2) edge (nb);

\draw (and5.input 1) edge (na);
\draw (and5.input 2) -- ++ (down:0.25) -| (or5);

\draw (or4.input 1) -- ++ (down:0.25) -| (a1);
\draw (or4.input 2) edge (na1);

\draw (or5.input 1) edge (b);
\draw (or5.input 2) -- ++ (down:0.25) -| (nb1);
\end{tikzpicture}
}
\end{subfigure}
\begin{subfigure}[b]{0.1\columnwidth}
\scalebox{.388}{
\begin{tikzpicture}[baseline=-300pt]
\node[circle,draw,minimum size=0.8cm] (a1) at (0,0.5) {};
\node[circle,draw,minimum size=0.8cm] (a2) at (0,1.5) {};
\node[circle,draw,minimum size=0.8cm] (a3) at (0,2.5) {};

\node[circle,draw,minimum size=0.8cm] (b1) at (1.5,0) {};
\node[circle,draw,minimum size=0.8cm] (b2) at (1.5,1) {};
\node[circle,draw,minimum size=0.8cm] (b3) at (1.5,2) {};
\node[circle,draw,minimum size=0.8cm] (b4) at (1.5,3) {};

\foreach \i in {1,...,3} {
    \foreach \j in {1,...,4} {
        \draw (a\i) -- (b\j);
    }
}

\node[circle,draw,minimum size=0.8cm] (c1) at (3.0,0) {};
\node[circle,draw,minimum size=0.8cm] (c2) at (3.0,1) {};
\node[circle,draw,minimum size=0.8cm] (c3) at (3.0,2) {};
\node[circle,draw,minimum size=0.8cm] (c4) at (3.0,3) {};

\foreach \i in {1,...,4} {
    \foreach \j in {1,...,4} {
        \draw (b\i) -- (c\j);
    }
}

\node[circle,draw,minimum size=0.8cm] (d1) at (4.5,0.5) {$\color{blue}0.3$};
\node[circle,draw,minimum size=0.8cm] (d2) at (4.5,1.5) {$\color{ForestGreen}0.5$};
\node[circle,draw,minimum size=0.8cm] (d3) at (4.5,2.5) {$\color{red}0.2$};

\foreach \i in {1,...,4} {
    \foreach \j in {1,...,3} {
        \draw (c\i) -- (d\j);
    }
}

\end{tikzpicture}
}
\end{subfigure}
\hfill
\begin{subfigure}[b]{0.45\columnwidth}
\centering
\hspace{-16pt}
\scalebox{0.7}{
\begin{tikzpicture}[circuit logic US]
\node (output) at (4.25, 8) {};
\node (or1) [or gate, inputs=nn, rotate=90, scale=0.9] at (4.25,7) {};
\node (or1val) [color=DarkOrchid] at ($(or1) + (0.7,0.1)$) {$\color{DarkOrchid}1.64$};
\draw (or1) -- (output);

\node (and1) [and gate, inputs=nn, rotate=90, scale=0.9] at (3,5.8) {};
\node (and1val) [color=DarkOrchid] at ($(and1) + (0.7,0.1)$) {$1.30$};
\node (and2) [and gate, inputs=nn, rotate=90, scale=0.9] at (5.7,5.8) {};
\node (and2val) [color=DarkOrchid] at ($(and2) + (0.7,0.1)$) {$1.04$};

\node (c) at (2.2,4.6) {$C$};
\node (cval) [color=DarkOrchid] at ($(c) - (0,0.4)$) {$0$};
\node (nc) at (7,4.6) {$\neg C$};
\node (ncval) [color=DarkOrchid] at ($(nc) - (0,0.4)$) {$0$};
\node (or2) [or gate, inputs=nnn, rotate=90, scale=0.9] at (3.1,4.6) {};
\node (or2val) [color=DarkOrchid] at ($(or2) + (0.7,0.1)$) {$1.30$};
\node (or3) [or gate, inputs=nn, rotate=90, scale=0.9] at (5.6,4.6) {};
\node (or3val) [color=DarkOrchid] at ($(or3) + (-0.7,0.1)$) {$1.04$};

\node (and3) [and gate, inputs=nn, rotate=90, scale=0.9] at (3.1,3.4) {};
\node (and3val) [color=DarkOrchid] at ($(and3) + (-0.7,0.1)$) {$1.30$};
\node (and4) [and gate, inputs=nn, rotate=90, scale=0.9] at (5.1,3.4) {};
\node (and4val) [color=DarkOrchid] at ($(and4) + (-0.6,0.1)$) {$0$};
\node (and5) [and gate, inputs=nn, rotate=90, scale=0.9] at (6.1,3.4) {};
\node (and5val) [color=DarkOrchid] at ($(and5) + (0.7,0.1)$) {$0.69$};

\node (a) at (5,1.7) {$A$};
\node (aval) [color=DarkOrchid] at ($(a) - (0,0.4)$) {$0$};
\node (nb) at (5.8,1.7) {$\neg B$};
\node (nbval) [color=DarkOrchid] at ($(nb) - (0,0.4)$) {$0$};
\node (na) at (6.6,1.7) {$\neg A$};
\node (naval) [color=DarkOrchid] at ($(na) - (0,0.4)$) {$0$};

\node (or4) [or gate, inputs=nn, rotate=90, scale=0.9] at (2.6,2.2) {};
\node (or4val) [color=DarkOrchid] at ($(or4) + (-0.7,0.1)$) {$0.61$};
\node (or5) [or gate, inputs=nn, rotate=90, scale=0.9] at (3.6,2.2) {};
\node (or5val) [color=DarkOrchid] at ($(or5) + (0.8,0.1)$) {$0.69$};

\node (a1) at (2.0,1) {$A$};
\node (a1val) [color=DarkOrchid] at ($(a1) - (0,0.4)$) {$0$};
\node (na1) at (2.7,1) {$\neg A$};
\node (na1val) [color=DarkOrchid] at ($(na1) - (0,0.4)$) {$0$};
\node (b) at (3.5,1) {$B$};
\node (bval) [color=DarkOrchid] at ($(b) - (0,0.4)$) {$0$};
\node (nb1) at (4.2,1) {$\neg B$};
\node (nb1val) [color=DarkOrchid] at ($(nb1) - (0,0.4)$) {$0$};

\draw (or1.input 1) -- ++ (down: 0.25) -| (and1);
\draw (or1.input 2) -- ++ (down: 0.25) -| (and2);

\draw (and1.input 1) -- ++ (down: 0.25) -| (c);
\draw (and1.input 2) -- (or2);

\draw (and2.input 2) -- ++ (down: 0.25) -| (nc);
\draw (and2.input 1) -- (or3);

\draw (or2.input 2) -- (and3);

\draw (or3.input 1) -- ++ (down: 0.25) -| (and4);
\draw (or3.input 2) -- ++ (down: 0.25) -| (and5);

\draw (and3.input 1) -- ++ (down: 0.25) -| (or4);
\draw (and3.input 2) -- ++ (down: 0.25) -| (or5);

\draw (and4.input 1) edge (a);
\draw (and4.input 2) -- ++ (down:0.8) -| (nb);

\draw (and5.input 1) -- ++ (down:0.55) -| (na);
\draw (and5.input 2) -- ++ (down:0.25) -| (or5);

\draw (or4.input 1) -- ++ (down:0.25) -| (a1);
\draw (or4.input 2) edge (na1);

\draw (or5.input 1) edge (b);
\draw (or5.input 2) -- ++ (down:0.25) -| (nb1);
\end{tikzpicture}
}
\end{subfigure}
\caption{
For a given data point, the network (middle) outputs a distribution over classes $A, B$ and $C$, highlighted in blue, green and red, respectively.
The circuit encodes the constraint $(a \land b) \implies c$.
For each leaf node $l$, we plug in $\Pr(l)$ and $1 - \Pr(l)$ for positive and negative literals, respectively.
The computation proceeds bottom-up, taking products at AND gates and summations at OR gates.
The value accumulated at the root of the circuit (left) is the probability allocated by the network to the constraint.
The weights accumulated on edges from OR gates to their children are of special significance: OR nodes induce a partitioning of the distribution's support, and the weights correspond to the mass allocated by the network to each mutually-exclusive event.
Complemented with a second upward pass, where the entropy of an OR node is the entropy of the distribution over it's children plus the expected entropy of its children, and the entropy of an AND node is the product of its children's entropies, we get the entropy of the distribution over the constraint's models -- the neuro-symbolic entropy regularization loss (right).
}
\label{fig:example}
\end{figure}

Lastly, a circuit is said to be  \emph{deterministic} if, for any input, at most one child of every OR node has a non-zero output i.e. for $\alpha = \bigvee_i \beta_i,\ \beta_i \land \beta_j = \bot \ \text{for} \  i \neq j$. Figure \ref{fig:example} shows an example of smooth, decomposable and deterministic circuit.

\subsection{Algorithm}
Let $\alpha$ be a \emph{smooth}, \emph{deterministic} and \emph{decomposable} logical circuit encoding our constraint, defined over Boolean variables $\Ys = \{Y_1,\dots,Y_n\}$. 
We now show that we can compute the constrained entropy in eqn~(\ref{eqn:nsentropy}) in time linear in the size $\alpha$.
The key insight is, using circuits, we're able to efficiently decompose an expectation with respect to a distribution by alternately splitting the query variables and the support of the distribution till we reach the leaves of the circuit -- literals -- when we proceed by combining solutions to our subproblems.

\subsubsection{Base Case: $\alpha$ is a literal}
When $\alpha$ is a literal, $l = y_i$ or $l = \lnot y_i$, we have that
\begin{align*}
    \Pr(y_i|\alpha) &= \Ind{y_i \models [\alpha]}, \text{~~and} \\
    H(y_i | \alpha)  &= - \Pr(y_i|\alpha) \log \Pr(y_i|\alpha) = 0.
\end{align*}
Intuitively, a literal has no uncertainty associated with it.
\subsubsection{Recursive Case: $\alpha$ is a conjunction}
When $\alpha$ is a conjunction, decomposability enables us to write
\begin{equation*}
    \Pr(\y|\alpha) = \Pr(\y_1|\beta) \Pr(\y_2|\gamma), \text{~where~}  \vars(\beta) \cap \vars(\gamma) = \varnothing
\end{equation*}
as it decomposes 
    $\alpha$ into two independent constraints $\beta$ and~$\gamma$,
    and $\y$ into two independent assignments $\y_1$ and~$\y_2$.
The neuro-symbolic entropy $- \mathbb{E}_{\Y | \alpha} \left[ \log \Pr(\Y | \alpha) \right]$ thus becomes
\begin{align*}
    &- \mathbb{E}_{\{\Y_1,\Y_2\} | \alpha} \Big[ \log \Pr(\Y_1|\beta) + \log \Pr(\Y_2|\gamma)\Big]\\
    &\quad =- \Big[\mathbb{E}_{\Y_1 | \beta} \big[ \log \Pr(\Y_1|\beta) \big] + \mathbb{E}_{\Y_2 | \gamma} \big [\log \Pr(\Y_2|\gamma)\big]\Big].
\end{align*}
That is, the entropy given a decomposable conjunction $\alpha$ is the sum of entropies given the conjuncts of~$\alpha$.
\subsubsection{Recursive Case: $\alpha$ is a disjunction}
When $\alpha$ is a smoothness and deterministic disjunction, 
we have that $\alpha = \bigvee_i \beta_i$, where the $\beta_i$s are mutually exclusive, and therefore partition $\alpha$. Consequently, we have that
\begin{equation*}
    \Pr(\y|\alpha) = \sum_i \Pr(\beta_i) \cdot \Pr(\y|\beta_i).
\end{equation*}
The neuro-symbolic entropy decomposes as well:
\begingroup
\allowdisplaybreaks
\begin{align*}
    &- \mathbb{E}_{\Y | \alpha} \left[ \log \Pr(\Y | \alpha) \right] 
    = -\sum_{\y \models \alpha}\Pr(\y|\alpha)\log \Pr(\y|\alpha)\\
    &= -\sum_{\y \models \alpha} \sum_i \Pr(\beta_i)\Pr(\y|\beta_i) \log \Big[\sum_j \Pr(\beta_j)\Pr(\y|\beta_j)\Big]\\
    \begin{split}
      &= -\sum_{\y \models \alpha} \sum_i \Pr(\beta_i)\Pr(\y|\beta_i)\id{\y \models \beta_i}\\ 
      &\qquad\qquad\qquad\log \Big[\sum_j \Pr(\beta_j)\Pr(\y|\beta_j)\id{\y \models \beta_j}\Big],
    \end{split}
    \intertext{where by determinism, we have that, for any $\y$ such that $\y \models \alpha$, $\y \models \beta_i \implies \y \not \models \beta_j$ for all $i \neq j$. In other words, any state that satisfies the constraint $\alpha$ satisfies one and only one of it's terms, and therefore, the above expression is equal to}
    &-\sum_{\y \models \alpha} \sum_i \Pr(\beta_i) \Pr(\y|\beta_i)\log \Big[\Pr(\beta_i)\Pr(\y|\beta_i)\Big]\id{\y \models \beta_i}\\
    &= -\sum_i \sum_{\y \models \beta_i} \Pr(\beta_i) \Pr(\y|\beta_i)\log \Big[\Pr(\beta_i)\Pr(\y|\beta_i)\Big].\\
    \intertext{Further simplifying the above expression, expanding the logarithm, and using the fact conditional probability sums to 1}
    \begin{split}
        &=-\sum_i \Pr(\beta_i) \log \Pr(\beta_i) \sum_{\y \models \beta_i} \Pr(\y|\beta_i)\\
        &\qquad\qquad\qquad+ \Pr(\beta_i) \sum_{\y \models \beta_i} \Pr(\y|\beta_i)\log \Pr(\y|\beta_i)
    \end{split}\\
    &=-\sum_i \Pr(\beta_i) \log \Pr(\beta_i) + \Pr(\beta_i) \mathbb{E}_{\Y | \beta_i} \Big[ \log \Pr(\Y|\beta_i) \Big].
\end{align*}
\endgroup
That is, the entropy of the random variable $\Y$ conditioned on a disjunction $\alpha$ is the sum of the entropy of the distributions induced on the children of $\alpha$, and the average entropy of its children. The full algorithm is illustrated in Algorithm \ref{alg:Shannon-Entropy}.

\section{Experimental Evaluation}\label{sec:experiments}
In this section we set out to empirically test our neuro-symbolic entropy loss.
To that end, we devise a series of semi-supervised and fully-supervised
structured prediction experiments.
Such are settings where, contrary to the their dominant use, classifiers are expected to predict structured
objects rather than scalar, discrete or real values. 
Such objects are defined in terms of constraints: a set of rules characterizing the set of solutions. 
We aim to answer the following:
\begin{enumerate}
    \item Does entropy regularization, in general, lead to predictive models 
    with improved generalization capabilities?
    \item If the answer to the above question is in the positive, it is our 
    expectation that restricting the distribution acted upon by entropy
    regularization to that over just the models of the constraint might
    seem more sensible as compared to entropy-regularizing the entire
    predictive distribution--including non-models of the constraint.
    Do experiments corroborate such a hypothesis?
    \item Finally, entropy regularization can be interpreted as clustering
    the different classes, and has intimate connections to transductive
    Support Vector Machines \cite{ssl}. Does such an interpretation carry
    over to models and non-models of the constraint? Put differently, can
    we expect entropy-regularized predictive models to better conform to
    our constraints, measured by the percentage of predictions satisfying
    the constraint \emph{regardless} of their correctness.
\end{enumerate}

\subsection{Semi-Supervised: Entity-Relation Extraction}

\newcommand{\tablescale}{0.97}

\begin{table*}[!htb]
\caption{Experimental results for joint entity-relation extraction on ACE05 and SciERC. \#Labels indicates the number of labeled data points made available to the network per relation. The remaining training set is stripped of labels and is utilized in an unsupervised manner.}
\centering
\scalebox{\tablescale}{%
\small
\begin{tabular}{llc|c|c|c|c|c|c}
\toprule
\# Labels  &    &3  &5  &10 &15 &25 &50 &75\\
\midrule
\multirow{6}{*}{\rotatebox[origin=c]{90}{ACE05}}
& Baseline
&4.92 $\pm$ 1.12          
&7.24 $\pm$ 1.75          
&13.66 $\pm$ 0.18 
&15.07 $\pm$ 1.79          
&21.65 $\pm$ 3.41          
&28.96 $\pm$ 0.98         
&33.02 $\pm$ 1.17 \\
& Self-training
&7.72 $\pm$ 1.21         
&12.83 $\pm$ 2.97            
&16.22 $\pm$ 3.08            
&17.55 $\pm$ 1.41            
&27.00 $\pm$ 3.66            
&32.90 $\pm$ 1.71           
&37.15 $\pm$ 1.42 \\
& Product t-norm
&8.89 $\pm$ 5.09       
&14.52 $\pm$ 2.13            
&19.22 $\pm$ 5.81            
&21.80 $\pm$ 7.67            
&30.15 $\pm$ 1.01           
&34.12 $\pm$ 2.75          
&37.35 $\pm$ 2.53 \\
\cmidrule{2-9}
& Semantic Loss
&12.00 $\pm$ 3.81
&14.92 $\pm$ 3.14 %
&22.23 $\pm$ 3.64
&27.35 $\pm$ 3.10
&30.78 $\pm$ 0.68
&36.76 $\pm$ 1.40
&38.49 $\pm$ 1.74\\
& + Full Entropy
&14.80 $\pm$ 3.70
&15.78 $\pm$ 1.90
&23.34 $\pm$ 4.07 %
&28.09 $\pm$ 1.46 %
&31.13 $\pm$ 2.26
&36.05 $\pm$ 1.00
&39.39 $\pm$ 1.21\\
& + NeSy Entropy
&14.72 $\pm$ 1.57
&18.38 $\pm$ 2.50
&26.41 $\pm$ 0.49
&31.17 $\pm$ 1.68
&35.85 $\pm$ 0.75
&37.62 $\pm$ 2.17
&41.28 $\pm$ 0.46\\
\midrule
\multirow{6}{*}{\rotatebox[origin=c]{90}{SciERC}}
& Baseline
&2.71 $\pm$ 1.1
&2.94 $\pm$ 1.0
&3.49 $\pm$ 1.8
&3.56 $\pm$ 1.1
&8.83 $\pm$ 1.0
&12.32 $\pm$ 3.0
&12.49 $\pm$ 2.6\\
&Self-training
&3.56 $\pm$ 1.4
&3.04 $\pm$ 0.9
&4.14 $\pm$ 2.6
&3.73 $\pm$ 1.1
&9.44 $\pm$ 3.8
&14.82 $\pm$ 1.2
&13.79 $\pm$ 3.9\\
&Product t-norm
&6.50 $\pm$ 2.0
&8.86 $\pm$ 1.2
&10.92 $\pm$ 1.6
&13.38 $\pm$ 0.7
&13.83 $\pm$ 2.9
&19.20 $\pm$ 1.7
&19.54 $\pm$ 1.7\\
\cmidrule{2-9}
&Semantic Loss
&$6.47\pm1.02$ 
&9.31 $\pm$ 0.76
&$11.50\pm1.53$
&$12.97\pm2.86$
&14.07 $\pm$ 2.33
&20.47 $\pm$ 2.50
&23.72 $\pm$ 0.38
\\
&+ Full Entropy
&$6.26\pm1.21$
&$8.49 \pm0.85$
&$11.12\pm1.22$
&14.10 $\pm$ 2.79 
&17.25 $\pm$ 2.75
&22.42 $\pm$ 0.43
&24.37 $\pm$ 1.62\\
&+ NeSy Entropy
&$6.19 \pm 2.40$
&$8.11 \pm 3.66$
&$13.17\pm1.08$
&$15.47\pm2.19$
&$17.45\pm1.52$
&$22.14\pm1.46$
&$25.11\pm1.03$\\
\bottomrule
\end{tabular}%
}
\label{table:results}
\end{table*}

We begin by testing our research questions in the semi-supervised setting.
Here the model is presented with only a portion of the labeled
training set, with the rest used exclusively in an unsupervised manner
by the respective approach.

We make use of the natural ontology of entity types and their relations 
present when dealing with relational data. This defines a set of 
relations and their permissible argument types. As is with all of our
constraints, we express the aforementioned ontology in the language 
of Boolean logic.

Our approach to recognizing the named entities and their pairwise relations
is most similar to \citet{Zhong2020}. Contextual embeddings are first procured
for every token in the sentence. These are then fed into a named entity 
recognition module that outputs a vector of per-class probability for every 
entity. A classifier then classifies the concatenated contextual 
embeddings and entity predictions into a relation.

We employ two entity-relation extraction datasets, the Automatic Content
Extraction (ACE) 2005 \cite{walker2006} and SciERC datasets \cite{luan2018}.
ACE05 defines an ontology over $7$ entities and $18$ relations from mixed-genre
text, whereas SciERC defines $6$ entity types with $7$ possible relation between
them and includes annotations for scientific entities and there relations,
assimilated from $12$ AI conference/workshop proceedings.
We report the percentage of coherent predictions: data points for which the
predicted entity types, as well as the relations are correct.

We compare against five baselines. The first baseline is a purely supervised
model which makes no use of unlabeled data. The second is a classical
self-training approach based off of \citet{chang2007}, and uses integer linear
programming to impute the unlabeled data's most likely labels subject to the
constraint, and consequently augment the (small) labeled set. The third 
baseline is a popular instantiation of a broad class of methods, fuzzy logics,
which replace logical operators with their fuzzy t-norms and logical implications
with simple inequalities. Lastly, we compare our proposed method, dubbed 
``NeSy Entropy'', to vanilla semantic loss as proposed in \citet{Xu18} 
as well as another entropy-regualrized baseline, dubbed ``Full Entropy'', which
minimizes the entropy of the entire predictive distribution, as opposed to just
the distribution over the constraint's models.

Our results are shown in Table \ref{table:results}. We observe that semantic loss outperforms
the baseline, self-training, and Product t-norm across the board. We attribute
such a performance to the exactness of semantic loss, and its faithfulness to
the underlying constraint. We also observe that entropy-regularizing the
predictive model, in conjunction with training using semantic loss leads to
better predictive models, as compared with models trained solely using semantic
loss. Furthermore, it turns out that restricting entropy to the distribution
over the constraint's models, models that we know constitute the set of valid
predictions, compared to the model's entire predictive distribution, which
includes valid and invalid predictions, leads to a non-trivial
increase in the accuracy of predictions.

\begingroup
\begin{table}[t]
\ra{1.05}
\centering
\caption {Test results for grids, preference learning, and warcraft}
\label{tab:gridres}
\scalebox{\tablescale}{\small
\begin{tabular}{ @{} l l c c c @{}}
\toprule
& Test accuracy \%  & Coherent & Incoherent & Constraint \\
\midrule
\multirow{4}{15pt}{\rotatebox[origin=c]{90}{Grid}} & 5-layer MLP & \phantom{0}5.6 & {\bf 85.9} & \phantom{0}7.0 \\
\cmidrule{2-5}
& Semantic loss & 28.5 & 83.1 & 69.9 \\
& + Full Entropy & 29.0 & 83.8  & 75.2 \\ 
& + NeSy Entropy & {\bf 30.1} & 83.0 & {\bf 91.6} \\
\midrule
\multirow{4}{15pt}{\rotatebox[origin=c]{90}{Preference}} & 3-layer MLP & \phantom{0}1.0 & {\bf 75.8} & \phantom{0}2.7 \\
\cmidrule{2-5}
& Semantic loss & 15.0 & 72.4 & 69.8 \\
& + Full Entropy & 17.5 & 71.8 & 80.2 \\
& + NeSy Entropy & {\bf 18.2} & 71.5 & {\bf 96.0} \\
\midrule
\multirow{4}{15pt}{\rotatebox[origin=c]{90}{Warcraft}} & ResNet-18&  44.8 & 97.7  & 56.9\\
\cmidrule{2-5}
& Semantic loss& 50.9 &  97.7 & 67.4\\
& + Full Entropy& 51.5 &  97.6&  67.7\\
& + NeSy Entropy & {\bf55.0}& {\bf97.9}& {\bf69.8}\\
\bottomrule
\end{tabular}
}
\label{tab:grid}
\label{tab:pref}
\label{tab:sp}
\end{table}

\subsection{Fully-Supervised Learning}
We now turn our attention to testing our hypotheses in a fully supervised setting,
where our aim is to examine the effect of constraints enforced on the training set.
We note that this is a seemingly harder setting in the following sense: In a semi-
supervised setting we might make the argument that, despite its abundance, imposing
an auxiliary loss on unlabeled data provides the predictive model with an unfair
advantage as compared to the baseline. We concern ourselves with two tasks: predicting paths in a grid and preference learning.

\paragraph{Predicting Simple Paths}
For this task, our aim is to find the shortest path in a graph, or more
specifically a 4-by-4 grid, $G = (V, E)$ with uniform edge weights. Our input is
a binary vector of length $|V| + |E|$, with the first $|V|$ variables indicating
the source and destination, and the next $|E|$ variables encoding a subgraph $G'
\subseteq G$. Each label is a binary vector of length $|E|$ encoding the
shortest \emph{simple} path in $G'$, a requirement that we enforce through our
constraint. We follow the algorithm proposed by \citet{nishino2017} to generate 
a constraint for each simple path in the grid, conjoined with indicators specifying
the corresponding source-destination pair. Our constraint is then the disjunction of all such conjunctions.

To generate the data, we begin by randomly removing one third of the edges in
the graph $G$, resulting in a subgraph, $G'$. Subsequently, we filter out
connected components in $G'$ with fewer than $5$ nodes to reduce degenerate
cases. We then sample a source and destination node uniformly at random. The
latter constitutes a single data point. We generate a dataset of $1600$
examples, with a $60/20/20$ train/validation/test split. %

\paragraph{Preference Learning}
We also consider the task of preference learning. Given the user's ranking of a subset of elements, we
wish to predict the user's preferences over the remaining elements of the set.
We encode an ordering over $n$ items as a binary matrix ${X_{ij}}$, where for
each $i, j \in {1, \ldots, n}$, $X_{ij}$ denotes that item $i$ is at position
$j$. Our constraint $\alpha$ requires that the network's output be a valid 
total ordering.
We use preference ranking data over $10$ types of sushi for $5,000$ individuals,
taken from PREFLIB \cite{MaWa13a}, split 60/20/20. Our inputs consist of the user's 
preference over $6$ sushi types, with the model tasked with predicting the
user's preference, a \emph{strict} total order, over the remaining $4$. %

Table \ref{tab:grid} compares the baseline
to the same MLP augmented with semantic loss, semantic
loss with entropy regularization over the entire predictive distribution, dubbed
``Full Entropy'' as well as entropy regularization over the distribution over the
constraint's models, dubbed ``NeSy Entropy".
Similar to \citet{Xu18}, we observe that the semantic loss has a marginal effect
on incoherent accuracy, but significantly improves the network’s  ability to output
coherent predictions. Furthermore, we observe that, similar to our semi-supervised
learning settings, entropy-regularization leads to more coherent predictions using
both ``Full Entropy'' and ``NeSy Entropy", with ``NeSy Entropy" leading to the
best performing predicting models. Remarkably, we also observe that ``NeSy Entropy''
leads to predictive models whose predictions almost always satisfy the constraint,
as captured by ``Constraint''.

\paragraph{Warcraft Shortest Path}
Lastly, we consider a more real-world variant of the task of predicting simple paths.
Following \citet{Pogancic2020}, our training set consists of $10,000$
terrain maps curated using Warcraft II tileset.
Each map encodes an underlying grid of dimension $12 \times 12$, where each vertex is assigned a cost depending on the type of terrain it represents (e.g. earth has lower cost than water).
The shortest (minimum cost) path between the top left and bottom right vertices is encoded as
an indicator matrix, and serves as label. 
\figref{fig:sp-results} shows an example input presented to the network, the groundtruth, and the input with the annotated shortest path.

\setlength{\fboxsep}{0pt}
\setlength{\picHeight}{0.25\linewidth}
\begin{figure}[t]
    \centering
        \scalebox{0.9}{
		\parbox[b][\picHeight][c]{1em}{\rotatebox{90}{Input}}
		\includegraphics[height=\picHeight]{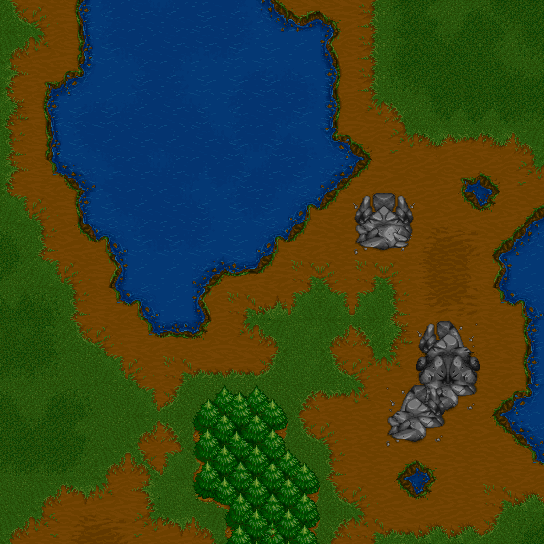}
		\vbox to\picHeight{\vfil\hbox{\LARGE$\to$}\vfil}
		\parbox[b][\picHeight][c]{\picHeight}{
		\centering
		$$
			\begin{pmatrix}
				1&0&\cdots&0\\
				1&0&\cdots&0\\
				\vdots & \vdots & \ddots &\vdots\\
				0&0&\cdots&1
			\end{pmatrix}
		$$} ~~~~~~
		\includegraphics[height=\picHeight]{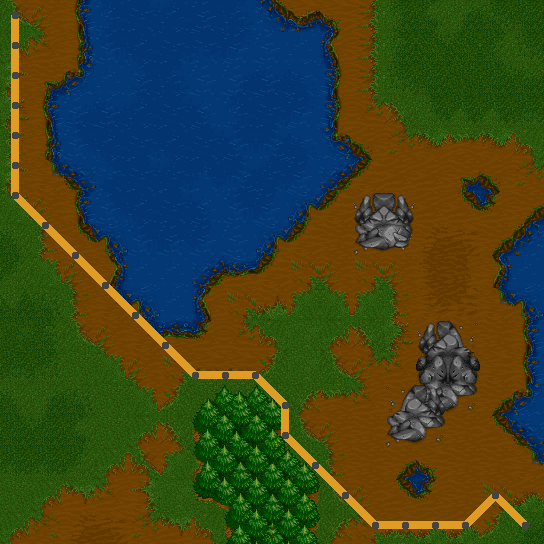}}
		\caption{Warcraft dataset. Each
		input (left) is a $12 \times 12$ grid corresponding to a Warcraft II
		terrain map, the output is a matrix (middle) indicating the shortest
		path from top left to bottom right (right).
		\label{fig:sp-results:dataset}
		\label{fig:sp-results:withpath}
		}
    \label{fig:sp-results}
\end{figure}

Presented with an image of a terrain map, a convolutional neural network -- following \citet{Pogancic2020}, we use ResNet18 \cite{He2016} -- outputs a $12 \times 12$ binary matrix indicating the vertices that constitute the minimum cost path.
We report three metrics: ``Coherent'' denotes the percentage of optimal-cost predictions, ``Incoherent'' denotes the percentage of individual vertices matching the groundtruth, and ``Constraint'' indicates the percentage of predictions that constitute valid paths. Our results are shown in Table \ref{tab:sp}.

In line with our previous experiments, we observe that incorporating constraints into learning improves the coherent accuracy from $44.8\%$ to $50.9\%$, and of valid predictions from $56.9\%$ to $67.4\%$.
Further augmenting semantic loss with the entropy over the network's predictive distribution, ``Full Entropy'', we attain a modest improvement from $50.9\%$ to $51.5\%$ and $67.4\%$ to $67.7\%$ for the ``Coherent'' and ``Constraint'' metrics respectively. 
Restricting the entropy minimization to models of the constraint, ``NeSy Entropy'', we observe that we attain a large improvement to $55.0\%$ and $69.8\%$ for the ``Coherent'' and ``Constraint'' metrics respectively.

\section{Related Work and Conclusion}\label{sec:related_work}
In an acknowledgment to the need for both symbolic as well as sub-symbolic reasoning, there has been a plethora of recent works studying how to best combine neural networks and logical reasoning, dubbed \emph{neuro-symbolic reasoning}. The focus of such approaches is typically making probabilistic reasoning tractable through first-order approximations, and differentiable, through reducing logical formulas into arithmetic objectives, replacing logical operators with their fuzzy t-norms, and logical implications with simple inequalities~\cite{rocktaschel2015,fischer19a}. 

\citet{diligenti2017} and \citet{donadello2017} use first-order logic to specify constraints on the outputs of a neural network. They employ fuzzy logic to reduce logical formulas into differential, arithmetic objectives denoting the extent to which neural network outputs violate the constraints, thereby supporting end-to-end learning under constraints. More recently, \citet{Xu18} introduced semantic loss, which circumvents the shortcomings of fuzzy approaches, while still supporting end-to-end learning under constraints. More precisely, \emph{fuzzy reasoning} is replaced with \emph{exact probabilistic reasoning}, made possible through compiling logical formulae into data structures that support efficient probabilistic queries.

Another class of neuro-symbolic approaches have their roots in logic programming. DeepProbLog~\cite{manhaeve2018} extends ProbLog, a probabilistic logic programming language, with the capacity to process neural predicates, whereby the network's outputs are construed as the probabilities of the corresponding predicates. This simple idea retains all essential components of ProbLog: the semantics, inference mechanism, and the implementation. In a similar vein, \citet{dai2018} combine domain knowledge specified as purely logical Prolog rules with the output of neural networks, dealing with the network's uncertainty through revising the hypothesis by iteratively replacing the output of the neural network with anonymous variables until a consistent hypothesis can be formed. \citet{bosnjak2017programming} present a framework combining prior procedural knowledge, as a Forth program, with neural functions learned through data. The resulting neural programs are consistent with the specified prior knowledge and optimized with respect to the data.

\label{sec:conclusion}
In conclusion, we proposed neuro-symbolic entropy regularization, a principled approach to unifying neuro-symbolic learning and entropy regularization. It encourages the network to output distributions that are peaked over models of the logical formula. We are able to compute our loss due to structural properties of circuit languages. We validate our hypothesis on four different tasks under semi-supervised and fully-supervised settings and observed an increase in \emph{accuracy} as well as the \emph{validity} of the predictions across the board.

\bibliographystyle{named}
\bibliography{references}

\end{document}